%% file: main.tex
\documentclass[10pt,twocolumn,letterpaper]{article}

\usepackage[pagenumbers]{cvpr} 


\usepackage{enumitem}
\usepackage{amsmath} 
\usepackage{amsfonts}
\usepackage{booktabs} 
\usepackage{multirow} 
\usepackage{xcolor}
\usepackage{colortbl}

\newlength\savewidth\newcommand\shline{\noalign{\global\savewidth\arrayrulewidth
		\global\arrayrulewidth 1pt}\hline\noalign{\global\arrayrulewidth\savewidth}}
\newcommand{\tablestyle}[2]{\setlength{\tabcolsep}{#1}\renewcommand{\arraystretch}{#2}\centering\footnotesize}
\renewcommand{\paragraph}[1]{\vspace{1.25mm}\noindent\textbf{#1}}
\newcommand\blfootnote[1]{\begingroup\renewcommand\thefootnote{}\footnote{#1}\addtocounter{footnote}{-1}\endgroup}

\newcolumntype{x}[1]{>{\centering\arraybackslash}p{#1pt}}
\newcolumntype{y}[1]{>{\raggedright\arraybackslash}p{#1pt}}
\newcolumntype{z}[1]{>{\raggedleft\arraybackslash}p{#1pt}}

\newcommand{\app}{\raise.17ex\hbox{$\scriptstyle\sim$}}

\definecolor{deemph}{gray}{0.6}

\definecolor{baselinecolor}{gray}{.9}

\definecolor{Gray}{gray}{0.9}

\usepackage{pifont}
\usepackage{xspace}
\newcommand{\one}{\ding{202}\xspace}
\newcommand{\two}{\ding{203}\xspace}
\newcommand{\three}{\ding{204}\xspace}

\definecolor{cvprblue}{rgb}{0.21,0.49,0.74}
\usepackage[pagebackref,breaklinks,colorlinks,allcolors=cvprblue]{hyperref}

\usepackage{marvosym}
\makeatletter
\def\blfootnote{\xdef\@thefnmark{}\@footnotetext}
\makeatother

\newcommand{\methodname}{FISA }

\def\paperID{} 
\def\confName{CVPR}
\def\confYear{2025}

\title{Adapting Vision-Language Model with Fine-grained Semantics for Open-Vocabulary Segmentation}

\author{Yong Xien Chng$^{1,2,*}$\quad Xuchong Qiu$^{2,\dagger}$\quad Yizeng Han$^{1}$\quad Kai Ding$^2$\quad Wan Ding$^2$\quad Gao Huang$^{1,}$\textsuperscript{\Letter}\\
$^1$Department of Automation, BNRist, Tsinghua University\quad $^2$Bosch Corporate Research\\
{\tt\small chngyx10@mails.tsinghua.edu.cn\quad gaohuang@tsinghua.edu.cn}
}

\begin{document}
\maketitle
\input{sec/0_abstract}    
\blfootnote{* Work done during internship at Bosch Corporate Research.}
\blfootnote{$\dagger$ Project lead.\quad \Letter\ Corresponding author.}

\input{sec/1_intro}
\input{sec/2_related}

\input{sec/3_analysis}

\input{sec/4_method}
\input{sec/5_experiment}
\input{sec/6_conclusion}

\noindent \textbf{Acknowledgements.} This work is supported in part by the National Natural Science Foundation of China under Grants 62321005 and 62276150, and the THU-Bosch JCML. The authors thank Lynn Tang and Qingyao Wang for their kind support in this project.

{
    \small
    \bibliographystyle{ieeenat_fullname}
    \bibliography{main}
}

\input{sec/X_suppl}

\end{document}

%% file: sec/0_abstract.tex
\begin{abstract}
Despite extensive research, open-vocabulary segmentation methods still struggle to generalize across diverse domains. To reduce the computational cost of adapting Vision-Language Models (VLMs) while preserving their pre-trained knowledge, most methods freeze the VLMs for mask classification and train only the mask generator. However, our comprehensive analysis reveals a surprising insight: open-vocabulary segmentation is primarily bottlenecked by mask classification, not mask generation. This discovery prompts us to rethink the existing paradigm and explore an alternative approach. Instead of freezing the VLM, we propose to freeze the pre-trained mask generator and focus on optimizing the mask classifier. Building on the observation that VLMs pre-trained on global-pooled image-text features often fail to capture fine-grained semantics necessary for effective mask classification, we propose a novel \underline{Fi}ne-grained \underline{S}emantic \underline{A}daptation (FISA) method to address this limitation. FISA enhances the extracted visual features with fine-grained semantic awareness by explicitly integrating this crucial semantic information early in the visual encoding process. As our method strategically optimizes only a small portion of the VLM's parameters, it enjoys the efficiency of adapting to new data distributions while largely preserving the valuable VLM pre-trained knowledge. Extensive ablation studies confirm the superiority of our approach. Notably, FISA achieves new state-of-the-art results across multiple representative benchmarks, improving performance by up to +1.0 PQ and +3.0 mIoU and reduces training costs by nearly 5$\times$ compared to previous best methods. Our code and data will be made public.

\end{abstract}

%% file: sec/1_intro.tex
\section{Introduction}

Open-vocabulary segmentation is an important task that \cite{panoptic,panopticdeeplab} combines semantic segmentation \cite{deeplab,fcn} of unseen background elements with instance segmentation \cite{maskrcnn} of unseen foreground objects. Its application has profound implications for enhancing scene comprehension in domains like autonomous driving \cite{codevilla2019exploring,toromanoff2020end} and robotics \cite{robot1,robot2}, leading to widespread research interest. Despite considerable progress, existing methods still show limited real-world performance and require substantial computational resources for training \cite{wu2024towards,zhu2023survey}, hindering their widespread adoption.

\begin{figure}[t]
  \centering
   \includegraphics[width=0.98\linewidth]{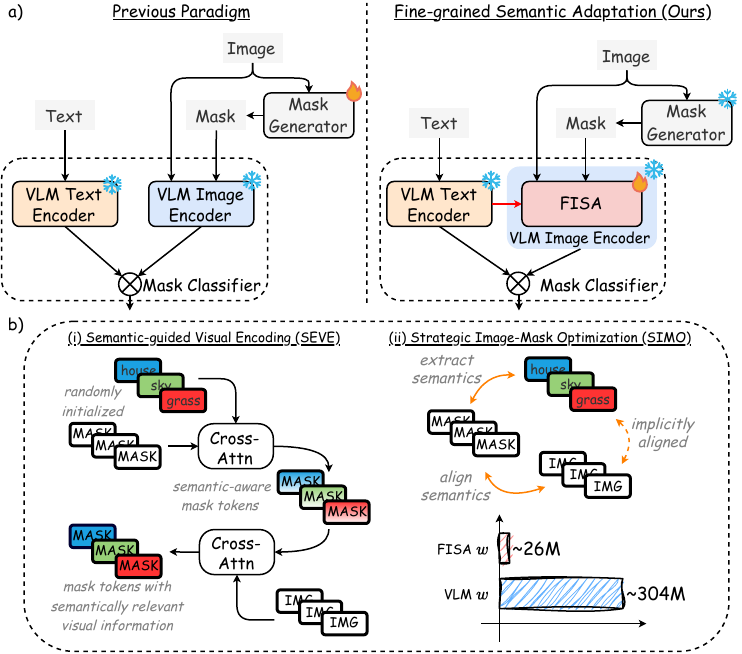}
   \captionsetup{aboveskip=5pt}
   \caption{a) Comparison between our proposed \underline{Fi}ne-grained \underline{S}emantic \underline{A}daptation (FISA) and previous open vocabulary segmentation paradigm. b) Unlike previous methods that focus on improving mask generation, \methodname adopts an alternative approach that focuses on improving mask classification. Specifically, it adopts a frozen pre-trained mask generator and enhances mask classification through two key innovations: i) Semantic-guided Visual Encoding that integrates fine-grained semantic information early in the visual encoding process, and ii) Strategic Image-Mask Optimization that selectively optimizes only a small portion of the VLM's parameters to retain its valuable pre-trained knowledge while endowing it with the flexibility to adapt to new distributions.}
   \label{figure1}

\end{figure}

Current open-vocabulary segmentation methods rely heavily on Vision-Language Models (VLMs) \cite{clip,openclip,EVA-CLIP} for their robust zero-shot capabilities \cite{xian2018zero}. These methods extract visual features from frozen VLMs and propose various techniques to utilize these features. They normally focus on training the mask generators and keep the VLMs frozen. The VLMs are kept frozen during training in order to minimize the high computational cost of adapting the large VLMs and to preserve their valuable pre-trained knowledge. However, since VLMs are generally not trained to process individual image regions, they may require some adaptation to perform optimally for dense segmentation tasks that require precise categorization of image parts. To verify this hypothesis, we conduct several analytical experiments using the highly modular and efficient MaskCLIP model \cite{maskclip}. As discussed in Sec.~\ref{section3}, our analysis reveals a surprising insight: \textit{mask classification is the primary performance bottleneck for open-vocabulary segmentation, and using an off-the-shelf pre-trained mask generator is already sufficient for this task}. In light of these observations, we decide to explore an alternative approach for open-vocabulary segmentation in this work. \textit{Instead of freezing the VLM, we freeze the pre-trained mask generator and focus exclusively on optimizing the VLM-based mask classifier}. To guide our strategy for improving mask classification, we conduct further investigation, which indicates that one of the main bottlenecks for mask classification stems from the \textit{lack of fine-grained semantic information in the visual features extracted by VLMs}. This suggests that enhancing the semantic awareness of these features could be a promising approach for improving mask classification performance.

Based on the insights gained from our preliminary analysis, we propose \underline{Fi}ne-grained \underline{S}emantic \underline{A}daptation (FISA), a novel framework for open-vocabulary segmentation that adopts a frozen pre-trained mask generator and fully focuses on improving mask classification by enhancing the fine-grained semantic richness of the extracted visual features through two key innovations. First, \methodname introduces a multimodal Semantic-guided Visual Encoding  mechanism (SEVE) that modifies CLIP's attention modules to infuse the relevant fine-grained semantic information early into the visual extraction process. This mechanism begins with mask tokens cross-attending with target class tokens, conditioning them on relevant semantic information. The semantically enriched mask tokens then cross-attend with image tokens to extract meaningful visual details, ultimately leading to more effective mask classification. Second, \methodname employs Strategic Image-Mask Optimization (SIMO) to selectively optimize only a small portion of the VLM’s parameters, preserving its valuable pre-trained knowledge while endowing it with the efficiency to adapt to new distributions.

Comprehensive experiments and ablations confirm the superiority of our method. Notably, \methodname achieves new state-of-the-art results across multiple key benchmarks and reduces training costs by nearly \textbf{5$\boldsymbol{\times}$} compared to the previous best method, MAFT+ \cite{maft+}. Specifically, \methodname outperforms MAFT+ by up to \textbf{1.0} points in PQ and \textbf{3.0} points in mIoU across multiple representative datasets. Our main contributions are summarized as follows:

\begin{enumerate}
\item We carefully analyze existing open-vocabulary segmentation methodology, revealing that \textit{mask classification is the main performance bottleneck for this task, and its weak performance mainly arises from a lack of fine-grained semantics in the extracted visual features}. 

\item We propose \underline{Fi}ne-grained \underline{S}emantic \underline{A}daptation (FISA), a novel framework that explores an alternative approach for open vocabulary segmentation. Contrary to existing methods that train their mask generators and freeze the VLMs, FISA utilizes a frozen pre-trained mask generator and effectively adapts the VLM to enrich its extracted visual features with fine-grained semantic information.

\item Despite using nearly \textbf{5$\boldsymbol{\times}$} less training cost than previous best method, our novel method sets new state-of-the-art results across multiple representative datasets. We extensively ablate our method to show its effectiveness.
\end{enumerate}

%% file: sec/2_related.tex
\section{Related Works}

\begin{figure*}[t]
    \centering
    \captionsetup{aboveskip=4pt}
    \begin{minipage}[b]{0.48\linewidth}
        \includegraphics[width=\linewidth]{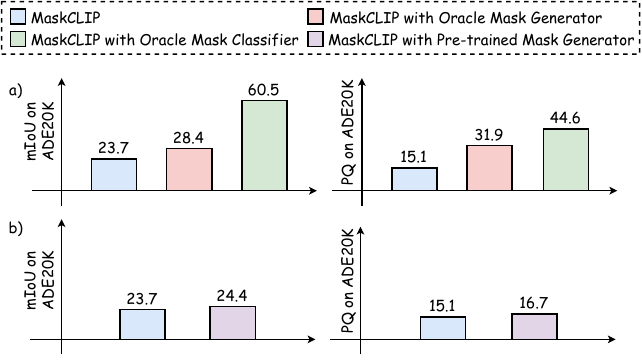}
        \captionsetup{aboveskip=5pt}
        \caption{a) MaskCLIP shows a much greater performance gain with a perfect ``oracle'' mask classifier than with a perfect ``oracle'' mask generator, highlighting \textit{mask classification as the main performance bottleneck for open-vocabulary segmentation}. b) Using a pre-trained mask generator performs as well as one re-trained from scratch, indicating that the mask generator can be frozen to enhance training efficiency without performance loss.}
        \label{figure2}
    \end{minipage}
    \hfill
    \begin{minipage}[b]{0.48\linewidth}
        \includegraphics[width=\linewidth]{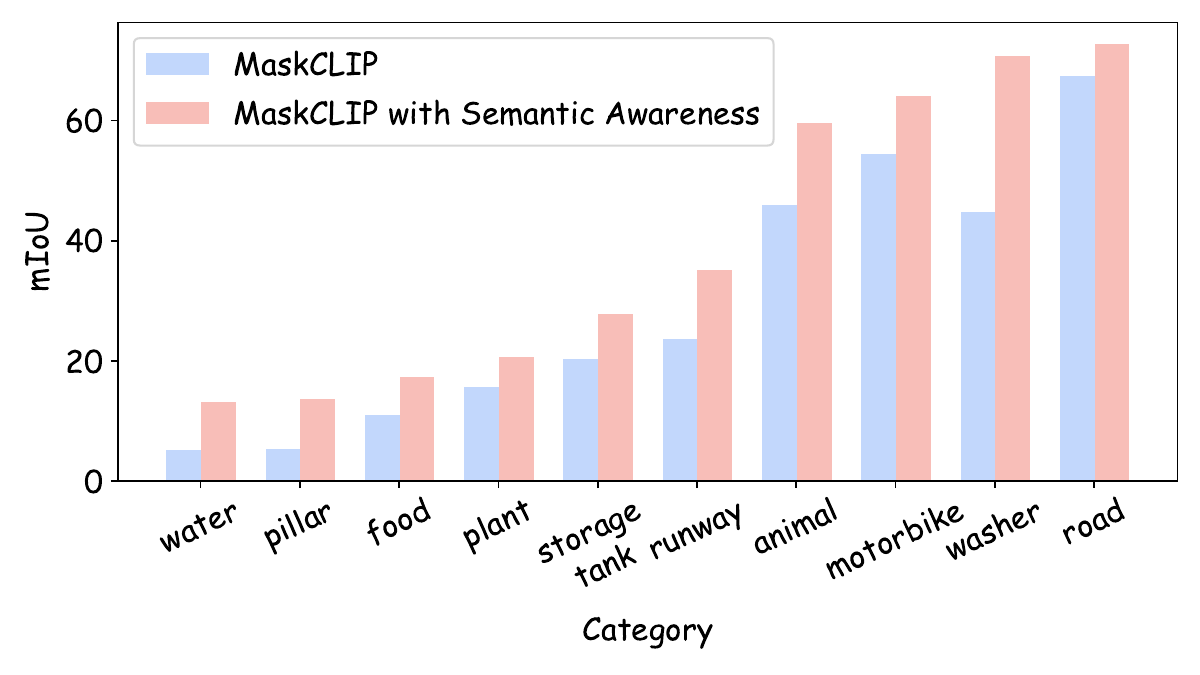}
        \captionsetup{aboveskip=5pt}
        \caption{The incorporation of fine-grained semantic awareness significantly improves MaskCLIP's performance across many out-of-domain classes in ADE20K. Compared to the baseline MaskCLIP model trained on COCO, this approach substantially improves performance, with gains of up to \textbf{13.7} points in mIoU. These results highlight the lack of fine-grained semantics as a key factor influencing performance in open-vocabulary segmentation.}
        \label{figure3}
    \end{minipage}
    \vskip -0.1in
\end{figure*}

\noindent \textbf{Open-vocabulary segmentation} combines both semantic and instance segmentation of unseen classes. Current methods primarily adopt Vision-Language Models (VLMs) such as CLIP \cite{clip,align,EVA-CLIP} that can perform zero-shot classification. Given the complexity of this task, research in this field begins with the exploration of methods focusing exclusively on open-vocabulary semantic segmentation. LSeg \cite{li2022languagedriven} directly fine-tunes a CLIP model to learn dense image features. While OpenSeg \cite{ghiasi2022scaling}, ZSseg \cite{xu2022simple}, and ZegFormer \cite{ding2022decoupling} all share a common approach of generating region proposals before applying CLIP classification, each implements this strategy differently. OVSeg \cite{liang2023open} collects mask-image pairs to improve CLIP's performance on masked images. SAN \cite{san} employs a side adapter network that leverages outputs from a frozen CLIP model to perform mask prediction and classification. CAT-Seg \cite{catseg} introduces a novel cost-aggregation method to refine CLIP's dense predictions. SED \cite{sed} further enhances CAT-Seg by using a hierarchical CLIP model to generate hierarchical dense predictions. As open-vocabulary semantic segmentation techniques mature and VLMs become increasingly sophisticated, attention shifts to the more challenging task of full open-vocabulary segmentation. Most methods for open-vocabulary segmentation initially adopt a two-stage approach for its simplicity and training efficiency. The pioneering MaskCLIP \cite{maskclip} introduces a novel Relative Mask Attention mechanism to extract regional mask information. MasQCLIP \cite{masqclip} enhances MaskCLIP by using progressive distillation to improve mask generation and adding a query adapter to enhance model adaptation. Since the two-stage approach generally lacks synergy between mask classification and generation, recent methods shift towards a one-stage approach to enhance performance. ODISE \cite{odise} explores using frozen internal representations of Stable Diffusion \cite{sd} for open-vocabulary panoptic segmentation, while FC-CLIP \cite{fcclip} investigates using a CNN-based CLIP model that efficiently provides feature maps with much higher resolution. To improve vision-text alignment, MAFT+ \cite{maft+} introduces a novel vision-text collaborative optimization to jointly optimize CLIP's vision and text representation. These methods generally freeze the VLMs used for mask classification and focus mainly on adapting the mask generators. However, in this work, we explore an alternative approach that adopts a frozen pre-trained mask generator and focuses exclusively on efficiently adapting the VLM-based mask classifier by enriching the extracted visual features with fine-grained semantic information.

\noindent \textbf{Efficient Adaptation Methods for VLMs} can significantly reduce the computational demands required for training these models. Among these approaches, adapter-based methods \cite{zhang2023llama,liu2024visual,gao2024clip} introduce minimal trainable parameters at strategic locations within the model, whereas prompt tuning \cite{jia2022visual, li2021prefix} injects these parameters into the input space. LoRA and its variants \cite{hu2021lora,dettmers2024qlora} avoid additional parameters by low-rank adapting only the linear layers. Alternatively, adapting the normalization layers \cite{zhao2023tuning} or the network biases \cite{cai2020tinytl} are also very effective in minimizing learnable parameters. In contrast to previous methods that entirely freeze their VLM-based mask classifiers, we explore fine-tuning a minimal subset of the VLM's parameters to improve its performance for open-vocabulary segmentation.

%% file: sec/3_analysis.tex
\section{Preliminary Analysis}
\label{section3}

\begin{figure*}[t]
   \vskip -0.1in
  \centering
   \includegraphics[width=\linewidth]{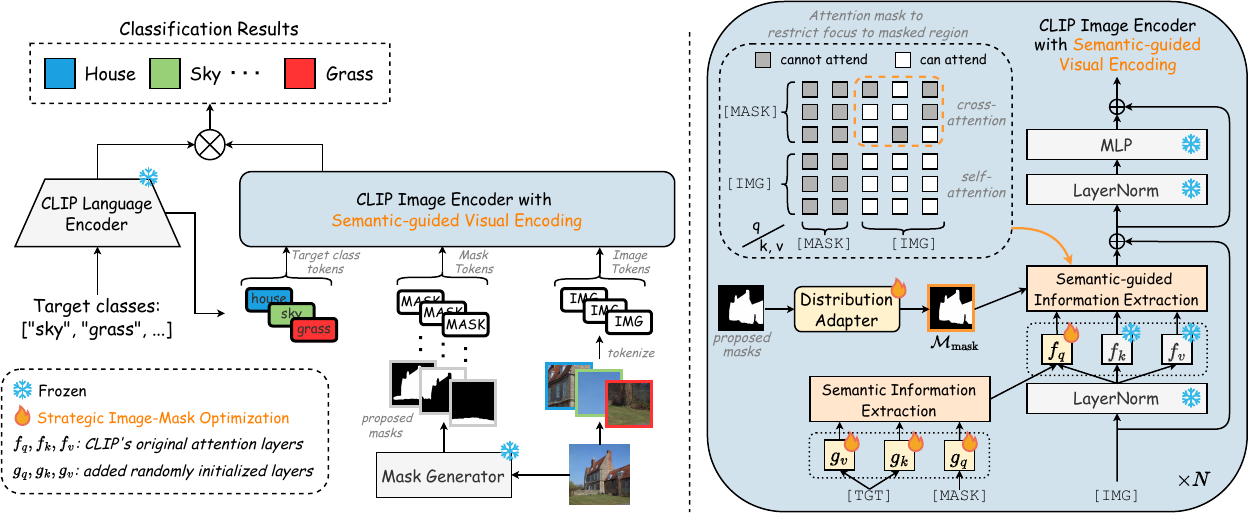}
   \captionsetup{aboveskip=5pt}
   \caption{Overview of \underline{Fi}ne-grained \underline{S}emantic \underline{A}daptation (FISA). Guided by the insight that \textit{mask classification is the main performance bottleneck and its weak performance mainly arises from the lack of of fine-grained semantics in the extracted visual features}, \methodname freezes the mask generator and introduces two key innovations for this task. First, it employs Semantic-guided Visual Encoding to inject semantic-awareness early into the visual feature extraction process. Second, it utilizes Strategic Image-Mask Optimization to efficiently adapt only a small number of CLIP's parameters to new data distributions while preserving its valuable pre-trained knowledge.}
   \label{figure4}
\end{figure*}

In this section, we carefully analyze the seminal MaskCLIP \cite{maskclip} method to identify key components affecting performance in open-vocabulary segmentation. This process yields critical insights that shape our approach in developing a method that performs effectively and trains efficiently. The results of our analysis are as follows:

\noindent \one \textbf{Between mask generation and mask classification, which step is the main performance bottleneck for open-vocabulary segmentation?} To answer this question, we conduct an experiment comparing the effects of a perfect mask generator and a perfect mask classifier on performance. We either replace the mask generator with an ``oracle" one that provides ground-truth masks, or replace the mask classifier with an ``oracle" one that assigns ground-truth labels on the predicted masks. Fig.~\ref{figure2}(a) shows that MaskCLIP with the ``oracle" classifier greatly outperforms MaskCLIP with the ``oracle" mask generator, achieving an mIoU of 60.5 and a PQ of 44.6 on the ADE150 dataset. This huge improvement of 32.1 points in mIoU and 12.7 points in PQ demonstrates that \textit{mask classification is the main performance bottleneck for open-vocabulary segmentation}.

\noindent \two \textbf{Can we improve training efficiency while maintaining model performance by freezing the pre-trained mask generator and focusing solely on mask classification?} To explore this possibility, we replace the mask generator with a COCO pre-trained version from Mask2Former’s model zoo \cite{mask2former}, keeping it frozen during training. As depicted in Fig.~\ref{figure2}(b), the performance of the pre-trained mask generator matches that of a newly trained one. This suggests that \textit{it is possible to freeze the mask generator, allowing us to focus solely on mask classification and enhance training efficiency without degrading performance.}

\noindent \three \textbf{What leads to the limited classification performance in existing open-vocabulary segmentation networks?} Upon examining existing networks, we observe that they primarily rely on mask attention \cite{maskclip} for extracting unimodal visual features while neglecting the fine-grained semantic information available in text labels during the visual feature extraction process. This omission is problematic because text information is crucial for aligning visual features with relevant semantic information. To validate this hypothesis, we compare MaskCLIP with our proposed variant that explicitly incorporates fine-grained semantics early into the feature extraction process. Specifically, we modify MaskCLIP to perform cross-attention with target-domain class labels before extracting visual features. As shown in Fig.~\ref{figure3}, this simple modification substantially improves performance across numerous out-of-domain classes in ADE20K, highlighting \textit{fine-grained semantic awareness as a crucial factor affecting network performance}.

\noindent \textbf{Summary}: Our analysis reveals a critical need to enhance mask classification performance in the development of open-vocabulary segmentation networks. Through examination of existing networks, we uncover a key limitation: the extracted visual features often lack fine-grained semantic details. This limitation arises from insufficient interaction between visual features and text labels during the feature extraction process and results in visual representations that often fail to capture the fine-grained semantic information essential for accurate segmentation. Building upon these insights, we now present our proposed solution.

%% file: sec/4_method.tex
\section{Method}
\label{section4}

In this section, we first describe the MaskCLIP \cite{maskclip} open-vocabulary segmentation framework, which our proposed \underline{Fi}ne-grained \underline{S}emantic \underline{A}daptation (FISA) is based upon. Following that, we explain in detail the core components of \methodname, namely 1) Semantic-guided Visual Encoding and 2) Strategic Image-Mask Optimization. Finally, we present the overall training loss function of our method.

\noindent \textbf{Architecture Overview.} As depicted in Fig.~\ref{figure4}, our method builds upon MaskCLIP \cite{maskclip}, which operates through sequential generation and classification of mask proposals. This process begins with a mask generator, which can be any conventional pre-trained segmentation network that is able to produce a set of candidate mask proposals. These proposals are then classified using a VLM capable of zero-shot classification. Following previous work, we utilize CLIP \cite{clip} for this purpose. CLIP consists of an image encoder and a language encoder. The image encoder extracts features from image tokens while the language encoder processes language labels. This model performs zero-shot classification by computing the cosine similarity between image and label embeddings, then assigning each image to the label with the highest similarity score. To adapt CLIP for the regional classification required by this task, we introduce a mask token for each mask proposal. These mask tokens attend only to image tokens within their corresponding masked regions. They function similarly as CLIP's $\mathtt{[CLS]}$ token by serving as compact vector representations of the information contained within each masked region.

\noindent \textbf{\underline{Fi}ne-grained \underline{S}emantic \underline{A}daptation (FISA)} is a simple yet effective framework that can substantially enhance the performance of existing open-vocabulary segmentation networks. Grounded in insights gained from previous analyses (Sec.~\ref{section3}), FISA freezes the mask generator and introduce two novel components to improve mask classification:
\begin{enumerate}
    \item \textbf{Semantic-guided Visual Encoding (SEVE).} Fine-grained semantic information is explicitly integrated into the visual feature encoding process, facilitating the extraction of semantically relevant feature representations.
    \item \textbf{Strategic Image-Mask Optimization (SIMO).} Only the additional parameters introduced for SEVE and the query projection layers within CLIP are updated, while all other layers remain frozen. This approach enables efficient cross-domain adaptation without compromising CLIP’s pre-trained knowledge.
\end{enumerate}

\noindent \textbf{Semantic-guided Visual Encoding (SEVE).} Current open-vocabulary segmentation methods typically rely on mask attention \cite{maskclip} for extracting regional information. However, this approach fails to leverage the semantic richness contained in text labels during visual feature extraction. This omission is problematic because textual information plays an essential role in aligning visual features with semantic content. To address this limitation, we propose SEVE, an innovative multimodal attention mechanism that directly integrates the relevant fine-grained semantic information early in the visual feature encoding process. Our approach involves two complementary steps. First, in the \textit{Semantic Information Extraction} step, the mask tokens cross-attend with target class tokens generated by CLIP's language encoder to infuse semantic understanding into the mask tokens, enabling them to capture contextually relevant information. Second, in the \textit{Semantic-guided Visual Information Extraction} step, these semantically-aware mask tokens cross-attend with tokens within the masked image regions to extract all task-specific and contextually relevant information. Before this cross-attention is applied, a lightweight Distribution Adapter, consisting of only two convolutional layers, is used to adjusts the mask proposals to align with CLIP's preferred input distribution. This adjustment is highly beneficial due to the large input-output distribution gap between the independently trained CLIP and the mask generator, as demonstrated in prior work \cite{liang2023open}. Mathematically, SEVE is computed as follows: Given $m$ mask tokens $\mathtt{[MASK]} \in \mathbb{R}^{m \times C}$, $n$ image tokens $\mathtt{[IMG]} \in \mathbb{R}^{n \times C}$,  $t$ target class tokens $\mathtt{[TGT]} \in \mathbb{R}^{t \times C}$, CLIP's query, key, value projections $f_q, f_k, f_v$, randomly initialized query, key, value projections $g_q, g_k, g_v$ for \textit{Semantic Information Extraction} and Softmax operator $\sigma$,
\small
\begin{gather}
\begin{aligned}
\hspace{-1.5mm} \text{SEVE}(\mathtt{[MASK]}, \mathtt{[IMG]},\mathtt{[TGT]}) = \sigma(\hat{\textbf{q}}_{\text{mask}}\textbf{k}_\text{img}^T + \mathcal{M}_{\text{mask}}) \cdot \textbf{v}_\text{img},
\end{aligned} \\
\hat{\textbf{q}}_\text{mask}, \textbf{k}_\text{img}, \textbf{v}_\text{img} = f_q(\hat{\mathtt{[MASK]}}),  f_k(\mathtt{[IMG]}), f_v(\mathtt{[IMG]}), \\
\hat{\mathtt{[MASK]}} = \sigma(\textbf{q}_{\text{mask}}\textbf{k}_\text{tgt}^T) \cdot \textbf{v}_\text{tgt}, \\
\textbf{q}_\text{mask}, \textbf{k}_\text{tgt}, \textbf{v}_\text{tgt} = g_q(\mathtt{[MASK]}),  g_k(\mathtt{[TGT]}), g_v(\mathtt{[TGT]}),
\end{gather}
\normalsize
where $\mathcal{M}_\text{mask} \in \mathbb{R}^{m \times n}$ is obtained by
\small
\begin{equation}
  \hspace{-2mm} \mathcal{M}_\text{mask}(i, j) =\begin{cases}
    0,&\hspace{-2mm} \text{if mask$_i$ contains any patch$_j$'s pixel},\\
    -\infty,&\hspace{-2mm} \text{otherwise}.
  \end{cases}
\end{equation}
\normalsize
\noindent Self-attention for image tokens is omitted here for brevity, as it remains unchanged from the original CLIP model.

\noindent \textbf{Strategic Image-Mask Optimization (SIMO).}  Although foundation models like CLIP already possess the necessary knowledge for open-vocabulary tasks, they often need some fine-tuning to adapt to new distributions. To principally guide our adaptation method, we revisit the Probably Approximately Correct (PAC) learning framework \cite{shalev2014understanding}. PAC explains a learning algorithm's generalization capability by relating it to the complexity of its hypothesis class $\mathcal{H}$ (i.e., the number of trainable parameters). Specifically, PAC connects the hypothesis class $\mathcal{H}$, a confidence level $\delta$, and a desired accuracy $\epsilon$ to determine the minimum sample size $m$ required for effective generalization, given by $m \geq \frac{\log(|\mathcal{H}|/\delta)}{\epsilon}$. This theorem suggests that when the sample size is fixed, reducing the model's parameters, thereby shrinking $\mathcal{H}$, can decrease the necessary sample size $m$ to achieve the same accuracy $\epsilon$ at the same confidence level $1 - \delta$. When applied to CLIP \cite{clip}, this principle highlights the importance of fine-tuning as few parameters as possible for effective cross-domain generalization. Based on this principle, we propose SIMO to optimize our network. SIMO strategically adapt only two set of parameters, namely those introduced for SEVE and the query projection layers within CLIP. All other layers in the mask classifier remain frozen. We show in Tab.~\ref{table3}(a) that our careful optimization approach is critical to enable efficient and effective adaptation to new domains, as both the fully fine-tuned or frozen CLIP perform much worse than our approach.

\noindent \textbf{Loss Function.} Following prior work \cite{maskclip}, we use a weighted combination of cross-entropy loss $\mathcal{L}_{\text{CE}}$, dice loss $\mathcal{L}_{\text{Dice}}$ and binary cross entropy loss $\mathcal{L}_{\text{BCE}}$ to train our model:
\begin{gather}
    \mathcal{L} = \lambda_{\text{CE}}\mathcal{L}_{\text{CE}} + \lambda_{\text{Dice}} \mathcal{L}_{\text{Dice}} + \lambda_{\text{BCE}}\mathcal{L}_{\text{BCE}}.
\end{gather}
In our experiments, we set $\lambda_{\text{CE}} = 2, \lambda_{\text{Dice}} = 5, \lambda_{\text{BCE}} = 5$.

%% file: sec/5_experiment.tex
\section{Experiments}
\label{section5}

\begin{table*}[t]
    \centering
    \resizebox{0.98\linewidth}{!}{
    \begin{tabular}{c|cc|cc|cc|c|c|c}
        \toprule
        \multirow{2}{*}{Method}   & \multicolumn{2}{c|}{COCO$^*$}  & \multicolumn{2}{c|}{ADE150} & \multicolumn{2}{c|}{Mapillary} & ADE847 & PC59 & PC459  \\
        &  PQ & mIoU & PQ & mIoU & PQ & mIoU & mIoU & mIoU & mIoU  \\
        \midrule
        OVSeg$^\dagger$ \cite{liang2023open} &  - & - & - & 29.6  & - & - & 9.0 & 57.7 & 15.7  \\
        SAN$^\dagger$ \cite{san}  &  - & - & -  & 33.3 & - & -  & 13.7 & 60.2& 17.1  \\
        SED$^\dagger$ \cite{sed}  & - & - & - & 35.2 & - & -  & 13.9 & 60.6 & 22.6 \\
        MaskCLIP \cite{maskclip} &  - & - & 15.1 & 23.7 & - & -  & 8.2 & 45.9 & 10.0  \\
        FreeSeg \cite{qin2023freeseg} &   - & - & 16.3 & 24.6 & - & -  & - & - & -\\
        ODISE \cite{odise} &   55.4 & 65.2& 22.6 & 29.9  & 14.2 & -  & 11.1 & 57.3 & 14.5  \\ 
        MasQCLIP \cite{masqclip}  & 48.5 & 62.0 & 23.3 & 30.4 & - & -  & 10.7 & 57.8 & 18.2  \\
        FC-CLIP \cite{fcclip} &    54.4 & 63.7 & 26.8 & 34.1 & 18.2 & 27.9  & 14.8 & 58.4 & 18.2  \\
        MAFT+ \cite{maft+} &  - & - & 27.1 & 36.1 & - & - & 15.1 & 59.4 & 21.6 \\
        \midrule
        \rowcolor{Gray} FISA (Ours) &  \textbf{56.4 $_{\text{\textcolor{ForestGreen}{(+2.0)}}}$} & \textbf{67.1 $_{\text{\textcolor{ForestGreen}{(+3.4)}}}$} & \textbf{28.1 $_{\text{\textcolor{ForestGreen}{(+1.0)}}}$} & \textbf{36.8 $_{\text{\textcolor{ForestGreen}{(+0.7)}}}$} & \textbf{19.0 $_{\text{\textcolor{ForestGreen}{(+0.8)}}}$} & \textbf{29.7 $_{\text{\textcolor{ForestGreen}{(+1.8)}}}$}  & \textbf{16.1 $_{\text{\textcolor{ForestGreen}{(+1.0)}}}$} & \textbf{62.4 $_{\text{\textcolor{ForestGreen}{(+3.0)}}}$} & \textbf{23.6 $_{\text{\textcolor{ForestGreen}{(+2.0)}}}$}  \\
        \bottomrule
    \end{tabular}
    }
    \captionsetup{aboveskip=5pt}
    \caption{Comparison with leading open-vocabulary panoptic segmentation and semantic segmentation methods. $^\dagger$ indicates models that can only perform semantic segmentation. $^*$ indicates close-vocabulary evaluation. \textbf{Bold} indicates best. }
    \label{table1}
\end{table*}

In this section, we first describe the datasets (Sec.~\ref{sec5.1}) and evaluation metrics (Sec.~\ref{sec5.2}) used. Next, we describe our implementation details (Sec.~\ref{sec5.3}). Then, we quantitatively and qualitatively compare our method with leading open-vocabulary segmentation methods (Sec.~\ref{sec5.4}). Finally, we carefully ablate our proposed method (Sec.~\ref{sec5.5}). 

\subsection{Training and Evaluation Datasets}
\label{sec5.1}
We train our method on the COCO-Panoptic \textit{train} dataset \cite{coco} and evaluate its performance using the COCO, Mapillary Vistas \cite{mapillary}, ADE20K \cite{zhou2019semantic} and PASCAL Context \cite{mottaghi_cvpr14} \textit{val} datasets. Splitting into \textit{train} and \textit{val} datasets with distinct labels is the standard practice in open-vocab. segmentation. Note that ADE20K has two subsets, ADE150 and ADE847, containing 150 and 847 classes, respectively. Similarly, PASCAL Context has two subsets, PC59 and PC459, containing 59 and 459 classes, respectively. 

\subsection{Evaluation Metrics}
\label{sec5.2}
We evaluate our method using two main metrics: Panoptic Quality (PQ) for panoptic segmentation and mean intersection-over-union (mIoU) for semantic segmentation. mIoU measures the average overlap between the predicted mask and the ground truth across all classes, while PQ measures the overall quality of a panoptic segmentation by combining semantic and instance segmentation accuracy. 

\subsection{Implementation Details}
\label{sec5.3}

We implement our method using Detectron2 \cite{wu2019detectron2} framework and follow the Mask R-CNN \cite{he2017mask} baseline settings 1 for training with COCO-Panoptic dataset. For our architecture, we employ the pre-trained ViT-L/16-336 CLIP model \cite{clip} as our mask classifier. Following FC-CLIP \cite{fcclip}, we use high-resolution input image with size $896 \times 896$. The position embeddings are adjusted through direct bilinear interpolation to accommodate the input size change, following standard practice in vision transformers \cite{li2022exploring,he2022masked,bolya2023window}. The text inputs to our model are the category names defined by each dataset. We use the pre-trained Swin-B Mask2Former segmentation model \cite{mask2former} as our mask generator without making any modification. We do not use Mask2Former's predicted class labels in our method. We train our model using the AdamW optimizer \cite{loshchilov2017decoupled} with a learning rate of $0.0001$, weight decay of 0.05, and a 0.1 learning rate multiplier for the feature backbone. Following MaskCLIP \cite{maskclip}, we employ large-scale jittering (LSJ) augmentation \cite{ghiasi2021simple} with random scale sampling from 0.1 to 2.0 and a fixed size crop to $1024 \times 1024$. The batch size is set to $16$, and the model is trained for $10,000$ iterations for all ablation experiments and $25,000$ iterations for the main results in Tab.~\ref{table1}. During inference, we follow standard Mask R-CNN settings, resizing images with shorter side to 800 and longer side up to 1333. For all other experimental settings not explicitly stated, we follow MaskCLIP's \cite{maskclip} settings.

\subsection{Main Results}
\label{sec5.4}

In this subsection, we quantitatively and qualitatively compare our method against the leading approaches using the COCO \cite{coco}, ADE20K \cite{zhou2019semantic}, and PASCAL \cite{mottaghi_cvpr14} datasets.

\noindent \textbf{Open-Vocabulary Panoptic Segmentation.} Tab.~\ref{table1} shows that our best method, FISA, outperforms both two-stage and one-stage approaches across various panoptic segmentation datasets. Compared to two-stage methods like MaskCLIP and MasQCLIP, FISA achieves a PQ improvement of up to \textbf{13.0} points on ADE150. Compared to one-stage methods like ODISE, FC-CLIP and MAFT+, FISA attains a PQ improvement of up to \textbf{5.5} points on both the indoor ADE150 and outdoor Mapillary Vistas datasets, establishing itself as the new state-of-the-art in this domain.

\noindent \textbf{Open-Vocabulary Semantic Segmentation.} Leading open-vocabulary semantic segmentation methods generally train on COCO-Stuff \cite{caesar2018coco} that provides extra annotations for semantic segmentation. Despite this unfair setup, our best method, FISA still manages to outperform all previous leading methods in semantic segmentation. Compared to the current best method, SED, FISA demonstrates improvements of \textbf{+1.6}, \textbf{+2.2}, \textbf{+1.8} and \textbf{+1.0} mIoU on ADE150, ADE847, PC59 and PC459, respectively. Furthermore, FISA also significantly outperforms all previous methods capable of performing both panoptic and semantic segmentation. Specifically, it surpasses the current leader in this category, MAFT+, by \textbf{+0.7}, \textbf{+1.0}, \textbf{+3.0} and \textbf{+2.0} mIoU on ADE150, ADE847, PC59, and PC459, respectively.

\noindent \textbf{Efficiency Analysis.} Tab.~\ref{table2} compares the efficiency of our method with several other leading open-vocabulary segmentation methods. FISA achieves competitive inference speed and significantly lower memory cost than the second-best method, MAFT+. Specifically, FISA requires only 45 GPU training hours and 10.2GB of GPU inference memory, compared to MAFT+'s 224 GPU hours and 13.7GB. This results in a substantial \textbf{5.0$\boldsymbol{\times}$} reduction in training time and \textbf{25.5$\boldsymbol{\%}$} memory savings. These efficiency gains arises from the considerable simplification of the segmentation pipeline and minimal parameters tuned (26M). Importantly, these efficiency improvements do not compromise performance, as our method continues to achieve state-of-the-art results across multiple representative datasets.

\begin{table}[ht]
    \centering
    \resizebox{\linewidth}{!}{\begin{tabular}{c | c c c  c | cc}
        \toprule
        \multirow{2}{*}{Method}  & Inference & Inference & Train & Train & \multicolumn{2}{c}{ADE150}  \\
        & FPS$\uparrow$ & Memory (GB)$\downarrow$ & GPU Hours$\downarrow$ & Iterations (K)$\downarrow$ & PQ$\uparrow$ & mIoU$\uparrow$ \\
        \midrule
        ODISE  &0.41 & - & 4760 & 369 &22.6 & 29.9 \\ 
        FC-CLIP  &  2.71 & 17.1 & 424 & 369 & 26.8 & 34.1  \\
        MAFT+  &  \textbf{2.94} & 13.7 & 224 & 60 & 27.1 & 36.1  \\
        \midrule
        \rowcolor{Gray} FISA (Ours) & 2.63 & \textbf{10.2} & \textbf{45} & \textbf{25} & \textbf{28.1}  & \textbf{36.8}  \\
        \bottomrule
    \end{tabular}}
    \captionsetup{aboveskip=5pt}
    \caption{Comparison with leading open-vocabulary segmentation methods on several important efficiency metrics.}
    \label{table2}
\end{table}

\begin{figure}[ht]
\centering
\includegraphics[width=0.95\linewidth]{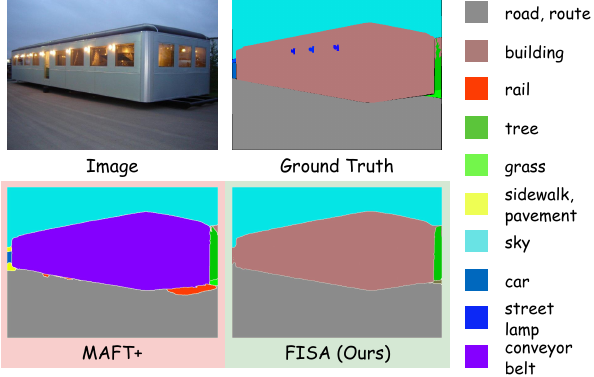}
\captionsetup{aboveskip=5pt}
\caption{Qualitative comparison on open-vocabulary semantic segmentation. Unlike MAFT+, our method accurately identifies buildings with uncommon shapes and textures while avoiding false predictions, such as misclassifying objects as \textit{rail}.} 
\label{smart_vis_semseg}
\end{figure}

\begin{figure}[ht]
\centering
\includegraphics[width=\linewidth]{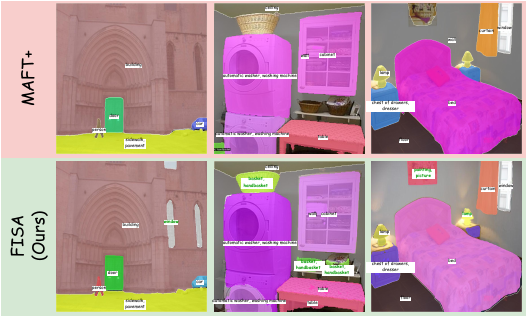}
\captionsetup{aboveskip=5pt}
\caption{Qualitative comparison on open-vocabulary panoptic segmentation. Unlike MAFT+, which often misses the predictions of certain objects, our method is able to produce more masks and achieves higher class prediction accuracy. Zoom-in for better view.} 
\label{figure6}
\end{figure}

\noindent \textbf{Qualitative Results.} In  Fig.~\ref{smart_vis_semseg} and Fig.~\ref{figure6}, we present some mask predictions of FISA on the ADE150 dataset for both semantic and panoptic segmentation. Compared to MAFT+, the current best method for open-vocab. segmentation, FISA generates more masks and predicts mask classes more accurately. 

\subsection{Ablations}
\label{sec5.5}

\begin{figure}[ht]
\centering
\includegraphics[width=\linewidth]{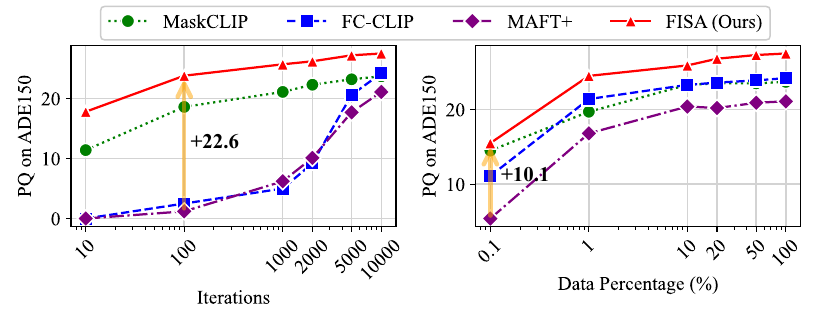}
\captionsetup{aboveskip=5pt}
\captionsetup{belowskip=-8pt}  
\caption{Effect of Training Length and Data Size on Model Performance. Our method consistently outperforms other leading methods across all training schedules and data sizes.} 
\label{figure7}
\end{figure}

\noindent \textbf{Robustness to Compute- and Data-Limited Scenarios.} Fig.~\ref{figure7} demonstrates FISA's robustness under limited training iterations and data sizes. As shown, our method consistently outperforms other leading methods under these settings. With just 100 training iterations, our method already outperforms the previous state-of-the-art approach, MAFT+, achieving a \textbf{+22.6} improvement on ADE150. Moreover, even when trained on a mere 0.1\% sample of the COCO-Panoptic dataset, our method still shows superior performance, surpassing MAFT+ by \textbf{+10.1} PQ on ADE150.

\begin{table*}[t]
\makebox[\textwidth][c]{\begin{minipage}{\linewidth}

\subfloat[
\textbf{Importance of Semantic-guided Visual Encoding (SEVE) and Strategic Image-Mask Optimization (SIMO).}
]{
\centering
\begin{minipage}{0.30\linewidth}{\begin{center}
\tablestyle{4pt}{1.05}
\hspace*{-2mm}
\begin{tabular}{lc}
Method & PQ \\
\shline
Baseline$^*$ & 9.6 \\
\quad + SEVE & 18.9 $_{\text{\textcolor{ForestGreen}{(+9.3)}}}$ \\
\quad \quad + SEVE + fully adapt CLIP & 14.9 $_{\text{\textcolor{ForestGreen}{(+5.3)}}}$ \\
\rowcolor{Gray} \quad \quad + SEVE + SIMO (FISA) & \textbf{\hspace{0.1mm} 27.5 $_{\text{\textcolor{ForestGreen}{(+17.9)}}}$} \\
\end{tabular}
\end{center}}\end{minipage}
}
\hspace{1em}
\subfloat[
\textbf{Comparison with LoRA \cite{hu2021lora}} FISA consistently outperforms LoRA across all ranks.
\label{tab:decoder_depth}
]{
\begin{minipage}{0.3\linewidth}{\begin{center}
\tablestyle{4pt}{1.05}
\begin{tabular}{lcc}
\\
Method & Rank & PQ \\
\shline
\rowcolor{Gray} FISA & - & \textbf{27.5} \\
LoRA & 256 & 25.3 $_{\text{\textcolor{BrickRed}{(-2.2)}}}$ \\
LoRA & 128 & 25.4 $_{\text{\textcolor{BrickRed}{(-2.1)}}}$ \\
LoRA & 64 & 25.9 $_{\text{\textcolor{BrickRed}{(-1.6)}}}$ \\
\end{tabular}
\end{center}}\end{minipage}
}
\hspace{1em}
\subfloat[
\textbf{Optimal size of Distribution Adapter.} Two layers provide the best performance.
]{
\begin{minipage}{0.30\linewidth}{\begin{center}
\tablestyle{4pt}{1.05}
\begin{tabular}{lx{36}x{24}}
\\
Number of layers & PQ \\
\shline
0 (w/o Distribution Adapter) & 26.7 \\
1 & 26.9 $_{\text{\textcolor{ForestGreen}{(+0.2)}}}$ \\
\rowcolor{Gray} 2 & \textbf{27.5 $_{\text{\textcolor{ForestGreen}{(+0.8)}}}$} \\
3 & 27.2 $_{\text{\textcolor{ForestGreen}{(+0.5)}}}$\\
\end{tabular}
\end{center}}\end{minipage}
}
\hspace{1em}
\\
\subfloat[
\textbf{Effect of adapting additional modules.} Adapting language encoder and mask generator do not provide performance gain.
]{
\begin{minipage}{0.3\linewidth}{\begin{center}
\tablestyle{4pt}{1.05}
\begin{tabular}{lc}
Parameters tuned & PQ \\
\shline
\rowcolor{Gray} FISA & \textbf{27.5} \\
\quad + adapt language encoder & 24.6 $_{\text{\textcolor{BrickRed}{(-2.9)}}}$ \\
\quad + adapt mask generator & 21.9 $_{\text{\textcolor{BrickRed}{(-5.6)}}}$ \\
\end{tabular}
\end{center}}\end{minipage}
}
\hspace{1em}
\subfloat[
\textbf{Scalability to VLM backbones.} FISA is compatible with different VLM backbones.
]{
\centering
\begin{minipage}{0.30\linewidth}{\begin{center}
\tablestyle{4pt}{1.05}
\begin{tabular}{lx{24}x{24}}
VLM backbone & PQ & mIoU \\
\shline
ViT-B-16 & 25.7 & 34.1 \\
\rowcolor{Gray} ViT-L-14-336  & \textbf{27.5} & \textbf{36.2} \\
EVA01-g-14-plus & 26.9 & 36.4 \\
\end{tabular}
\end{center}}\end{minipage}
}
\hspace{1em}
\subfloat[
\textbf{CLIP's oIoU performance on ref. segmentation before and after performing FISA.} 
]{
\begin{minipage}{0.30\linewidth}{\begin{center}
\tablestyle{4pt}{1.05}
\begin{tabular}{lcc}
\\
Case & RefCOCO & RefCOCO+ \\
\shline
without FISA & 23.9 & 25.0 \\
\rowcolor{Gray} with FISA & \textbf{24.6 $_{\text{\textcolor{ForestGreen}{(+0.7)}}}$} & \textbf{25.9 $_{\text{\textcolor{ForestGreen}{(+0.9)}}}$} \\
\\
\end{tabular}
\end{center}}\end{minipage}
}
\hspace{1em}
\end{minipage}}

\caption{\textbf{Ablation experiments} on ADE150 using FISA. All experiments here are run with a shorter training schedule of 10000 iterations, causing the results to be different from Tab.~\ref{table1}. The entries marked in {\setlength{\fboxsep}{0pt}\colorbox{baselinecolor}{gray}} are the same, which specify the default settings. $^*$Baseline is a direct combination of a frozen pre-trained mask generator and a frozen CLIP. }
\label{table3}
\end{table*}

\paragraph{Benefits of Semantic-guided Visual Encoding (SEVE) and Strategic Image-Mask Optimization (SIMO).} In Tab.~\ref{table3}(a), we incrementally integrate our proposed modules into the baseline model, which initially combines a frozen pre-trained mask generator with a frozen CLIP using mask attention to extract regional information \cite{maskclip}. Our proposed SEVE module significantly enhances this baseline, achieving a large improvement of +9.3 PQ. Additionally, when combined with SIMO, the performance further improves by +8.6 PQ, resulting in a final model that reaches 27.5 PQ, achieving a new state-of-the-art for this task. SIMO is crucial as fully adapting CLIP performs much worse.

\paragraph{Effect of Adapting Other Modules.} To efficiently preserve CLIP's pre-trained knowledge while giving it the flexibility to adapt to new distributions, we utilize a minimal adaptation approach called SIMO that selectively optimizes only a small subset of CLIP's parameters. In Tab.~\ref{table3}(b), we ablate the effect of tuning additional pre-trained modules, namely the mask generator and language encoder. The results reveal a sharp performance decline after these adjustments. This outcome validates the benefits of our optimization strategy and underscores the importance of careful parameter adjustment for cross-domain generalization.

\paragraph{Scalability to Different VLM Backbones.} In Tab. \ref{table3}(c), we investigate the scalability of our method with respect to various VLM backbones. As shown, FISA is compatible with different VLM backbones, allowing it to benefit easily from future advancements in VLM backbones.

\paragraph{Comparison with Other Efficient Fine-Tuning Methods.} Low-Rank Adaptation (LoRA) \cite{hu2021lora} is widely used for efficiently fine-tuning pre-trained networks in transfer learning. In Tab.~\ref{table3}(d), we compare FISA with LoRA. Following common practice, we apply LoRA to the attention projection layers of the CLIP model \cite{clip}. As shown in the table, our method significantly outperforms LoRA across different ranks, demonstrating its effectiveness in fine-tuning CLIP for open-vocabulary segmentation.

\paragraph{Effect of Using Different Number of Layers in Distribution Adapter.} In Tab.~\ref{table3}(e), we evaluate the sensitivity of Distribution Adapter in SEVE to different number of convolutional layers. We observe that using two layer achieves the best performance and adhere to this design choice.

\paragraph{Preservation of CLIP's Pre-trained Knowledge.} To show that CLIP's internal knowledge is preserved after applying FISA, we compare the original CLIP backbone's performance with our minimally adapted version on referring image segmentation \cite{chng2023mask}. As shown in Tab.~\ref{table3}(f), performance remains unchanged after adaptation. This is possible because of FISA's minimal adaptation approach, which restricts weight updates to a select few strategically chosen parameters and requires only a very small tuning iterations.

\paragraph{Compatibility with Mask Generators.} To demonstrate our method's compatibility with various mask generators, we conduct ablation studies using different pre-trained mask generators. As shown in Tab.~\ref{table4}, all tested mask generators produce meaningful results. While stronger mask generators show slight improvements, the overall performance gains are minimal. This observation further confirms that mask classification is the main bottleneck for this task, validating our approach of focusing on this aspect.  

\begin{table}[ht]
\centering
\resizebox{\linewidth}{!}{\begin{tabular}{c|cc|c|c|c}
    \toprule
    \multirow{2}{*}{Mask Generator Backbone} & \multicolumn{2}{c|}{ADE150} & ADE847 & PC59 & PC459 \\
    & PQ & mIoU & mIoU & mIoU & mIoU \\
    \midrule
    ResNet-50 & 26.1 & 35.9 & 15.8 & 61.5 & 22.9 \\
    Swin-T & 26.1 & 36.6 & 16.1 & 61.7 & 23.2 \\
    \rowcolor{Gray} Swin-B & \textbf{28.1} & \textbf{36.8} & \textbf{16.1} & \textbf{62.4} & \textbf{23.6} \\
    \bottomrule
\end{tabular}
}
\captionsetup{aboveskip=5pt}  
\captionsetup{belowskip=-5pt}  
\caption{Compatibility with different mask generators .}
\label{table4}
\end{table}  

%% file: sec/6_conclusion.tex
\section{Conclusion}
\label{section6}
In this paper, we rethink the existing paradigm for open-vocabulary segmentation and propose \underline{Fi}ne-grained \underline{S}emantic \underline{A}daptation (FISA), a novel framework that freezes the mask generator and efficiently adapts the VLM-based mask classifier to improve open-vocabulary segmentation performance. This exploration is grounded in the insight that \textit{mask classification is the main performance bottleneck for open-vocabulary segmentation and using an off-the-shelf mask generator is already sufficient for this task}. To guide our improvement strategy for mask classification, we analyze existing networks and find that their weak classification performance mainly stems from a \textit{lack of fine-grained semantics in the extracted visual features}. To address this limitation, FISA introduces two key innovations: 1) Semantic-guided Visual Encoding mechanism to inject fine-grained semantic awareness early into the visual feature encoding process, and 2) Strategic Image-Mask Optimization to optimize only a small subset of the VLM's parameters, providing the VLM with the flexibility to adapt to new data distributions while largely preserving its valuable pre-trained knowledge. Remarkably, our method achieves new state-of-the-art results across multiple key datasets while reducing training costs by nearly \textbf{5$\boldsymbol{\times}$} compared to the previous best method, MAFT+. 

\newpage

%% file: sec/X_suppl.tex
\clearpage
\setcounter{page}{1}
\onecolumn
\begin{center}
\Large
\textbf{\thetitle}\\
\vspace{0.5em}Supplementary Material \\
\vspace{1.0em}
\end{center}

\normalsize

\section{Additional Visualizations}

\begin{figure}[h]
  \centering
   \includegraphics[width=0.92\linewidth]{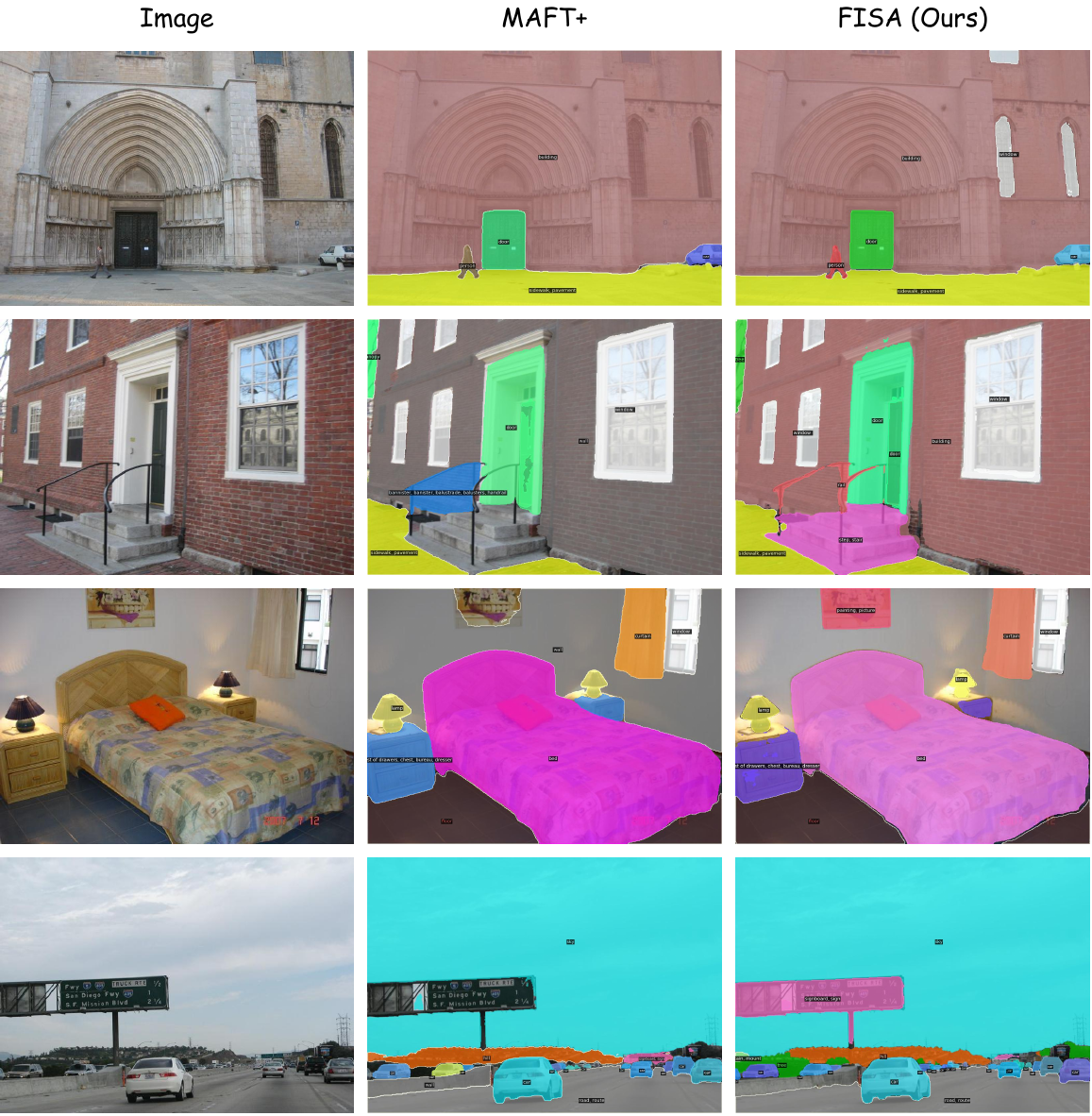}
   \caption{Visualizations of open-vocabulary segmentation predictions by our proposed FISA and previous best method, MAFT+ on the ADE150 validation dataset (1/3). Zoom in for best view.}
   \label{fig1}
\end{figure}

\begin{figure*}[h]
  \centering
   \includegraphics[width=\linewidth]{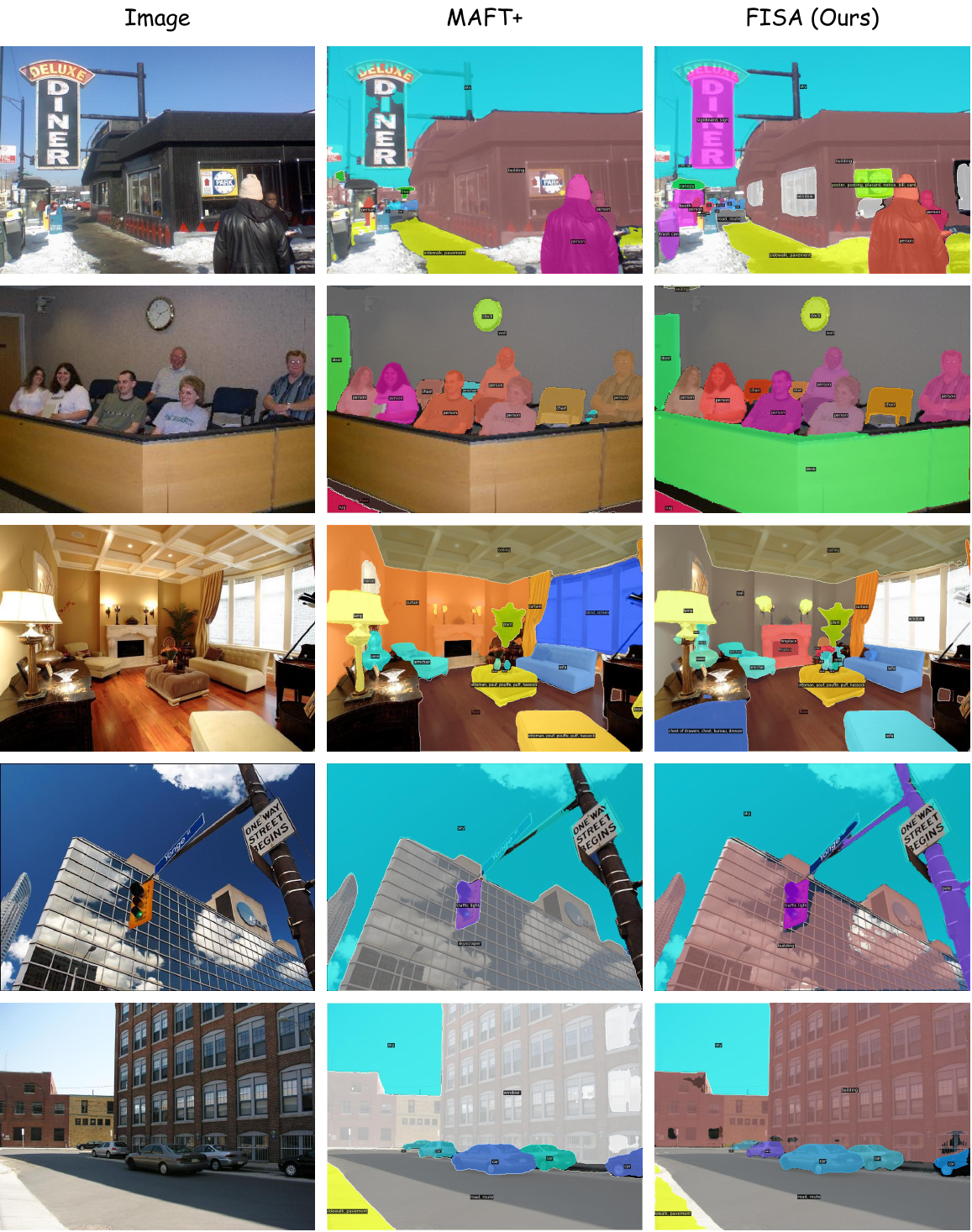}
   \caption{Visualizations of open-vocabulary segmentation predictions by our proposed FISA and previous best method, MAFT+ on the ADE150 validation dataset (2/3). Zoom in for best view.}
   \label{fig2}
\end{figure*}

\begin{figure*}[h]
  \centering
   \includegraphics[width=\linewidth]{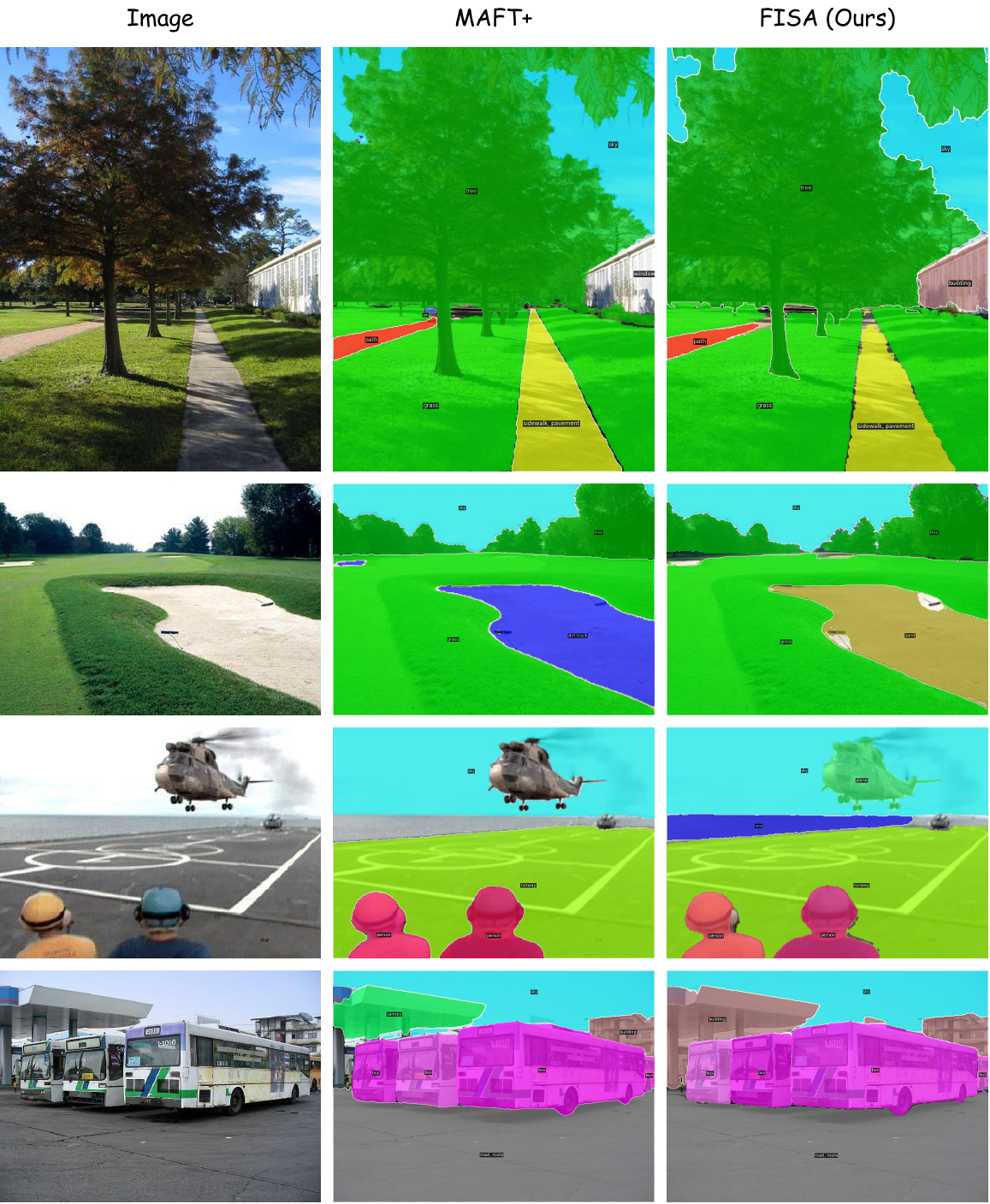}
   \caption{Visualizations of open-vocabulary segmentation predictions by our proposed FISA and previous best method, MAFT+ on the ADE150 validation dataset (3/3). Zoom in for best view.}
   \label{fig3}
\end{figure*}

\noindent In Fig.~\ref{fig1}, Fig.~\ref{fig2}, and Fig.~\ref{fig3}, we present additional qualitative comparisons between open-vocabulary segmentation predictions made by our proposed FISA method and the previous state-of-the-art method, MAFT+, on the ADE150 dataset.

%% file: main.bbl
\begin{thebibliography}{55}
\providecommand{\natexlab}[1]{#1}
\providecommand{\url}[1]{\texttt{#1}}
\expandafter\ifx\csname urlstyle\endcsname\relax
  \providecommand{\doi}[1]{doi: #1}\else
  \providecommand{\doi}{doi: \begingroup \urlstyle{rm}\Url}\fi

\bibitem[Ahn et~al.(2018)Ahn, Choi, Kim, Cha, and Oh]{robot2}
H. Ahn, S. Choi, N. Kim, G. Cha, and S. Oh.
\newblock Interactive text2pickup networks for natural language-based human--robot collaboration.
\newblock \emph{IEEE Robotics and Automation Letters}, 2018.

\bibitem[Bolya et~al.(2024)Bolya, Ryali, Hoffman, and Feichtenhofer]{bolya2023window}
D. Bolya, C. Ryali, J. Hoffman, and C. Feichtenhofer.
\newblock Window attention is bugged: How not to interpolate position embeddings.
\newblock In \emph{ICLR}, 2024.

\bibitem[Caesar et~al.(2018)Caesar, Uijlings, and Ferrari]{caesar2018coco}
Holger Caesar, Jasper Uijlings, and Vittorio Ferrari.
\newblock Coco-stuff: Thing and stuff classes in context.
\newblock In \emph{CVPR}, 2018.

\bibitem[Cai et~al.(2020)Cai, Gan, Zhu, and Han]{cai2020tinytl}
H. Cai, C. Gan, L. Zhu, and S. Han.
\newblock Tinytl: Reduce activations, not trainable parameters for efficient on-device learning.
\newblock In \emph{NeurIPS}, 2020.

\bibitem[Chen et~al.(2017)Chen, Papandreou, Kokkinos, Murphy, and Yuille]{deeplab}
L. Chen, G. Papandreou, I. Kokkinos, K. Murphy, and A. Yuille.
\newblock Deeplab: Semantic image segmentation with deep convolutional nets, atrous convolution, and fully connected crfs.
\newblock \emph{PAMI}, 2017.

\bibitem[Cheng et~al.(2020)Cheng, Collins, Zhu, Liu, Huang, Hartwig, and Chen]{panopticdeeplab}
B. Cheng, M. Collins, Y. Zhu, T. Liu, T. Huang, A. Hartwig, and L. Chen.
\newblock Panoptic-deeplab: A simple, strong, and fast baseline for bottom-up panoptic segmentation.
\newblock In \emph{CVPR}, 2020.

\bibitem[Cheng et~al.(2022)Cheng, Misra, Schwing, Kirillov, and Girdhar]{mask2former}
B. Cheng, I. Misra, A.~G. Schwing, A. Kirillov, and R. Girdhar.
\newblock Masked-attention mask transformer for universal image segmentation.
\newblock In \emph{CVPR}, 2022.

\bibitem[Cherti et~al.(2023)Cherti, Beaumont, Wightman, Wortsman, Ilharco, Gordon, Schuhmann, Schmidt, and Jitsev]{openclip}
M. Cherti, R. Beaumont, R. Wightman, M. Wortsman, G. Ilharco, C. Gordon, C. Schuhmann, L. Schmidt, and J. Jitsev.
\newblock Reproducible scaling laws for contrastive language-image learning.
\newblock In \emph{CVPR}, 2023.

\bibitem[Chng et~al.(2024)Chng, Zheng, Y.~Han, and Huang]{chng2023mask}
Y.~X. Chng, H. Zheng, X.~Qiu Y.~Han, and G. Huang.
\newblock Mask grounding for referring image segmentation.
\newblock In \emph{CVPR}, 2024.

\bibitem[Cho et~al.(2024)Cho, Shin, Hong, An, Lee, Arnab, Seo, and Kim]{catseg}
S. Cho, H. Shin, S. Hong, S. An, S. Lee, A. Arnab, P.~H. Seo, and S. Kim.
\newblock Cat-seg: Cost aggregation for open-vocabulary semantic segmentation.
\newblock In \emph{CVPR}, 2024.

\bibitem[Codevilla et~al.(2019)Codevilla, Santana, L{\'o}pez, and Gaidon]{codevilla2019exploring}
F. Codevilla, E. Santana, A.~M. L{\'o}pez, and A. Gaidon.
\newblock Exploring the limitations of behavior cloning for autonomous driving.
\newblock In \emph{ICCV}, 2019.

\bibitem[Dettmers et~al.(2024)Dettmers, Pagnoni, Holtzman, and Zettlemoyer]{dettmers2024qlora}
T. Dettmers, A. Pagnoni, A. Holtzman, and L. Zettlemoyer.
\newblock Qlora: Efficient finetuning of quantized llms.
\newblock In \emph{NeurIPS}, 2024.

\bibitem[Ding et~al.(2022)Ding, Xue, Xia, and Dai]{ding2022decoupling}
J. Ding, N. Xue, G.-S. Xia, and D. Dai.
\newblock Decoupling zero-shot semantic segmentation.
\newblock In \emph{CVPR}, 2022.

\bibitem[Ding et~al.(2023)Ding, Wang, and Tu]{maskclip}
Z. Ding, J. Wang, and Z. Tu.
\newblock Open-vocabulary universal image segmentation with maskclip.
\newblock In \emph{ICML}, 2023.

\bibitem[Gao et~al.(2024)Gao, Geng, Zhang, Ma, Fang, Zhang, Li, and Qiao]{gao2024clip}
P. Gao, S. Geng, R. Zhang, T. Ma, R. Fang, Y. Zhang, H. Li, and Y. Qiao.
\newblock Clip-adapter: Better vision-language models with feature adapters.
\newblock \emph{IJCV}, 2024.

\bibitem[Ghiasi et~al.(2021)Ghiasi, Cui, Srinivas, Qian, Lin, Cubuk, Le, and Zoph]{ghiasi2021simple}
Golnaz Ghiasi, Yin Cui, Aravind Srinivas, Rui Qian, Tsung-Yi Lin, Ekin~D Cubuk, Quoc~V Le, and Barret Zoph.
\newblock Simple copy-paste is a strong data augmentation method for instance segmentation.
\newblock In \emph{CVPR}, 2021.

\bibitem[Ghiasi et~al.(2022)Ghiasi, Gu, Cui, and Lin]{ghiasi2022scaling}
G. Ghiasi, X. Gu, Y. Cui, and T.-Y. Lin.
\newblock Scaling open-vocabulary image segmentation with image-level labels.
\newblock In \emph{ECCV}, 2022.

\bibitem[He et~al.(2017{\natexlab{a}})He, Gkioxari, Doll{\'a}r, and Girshick]{he2017mask}
Kaiming He, Georgia Gkioxari, Piotr Doll{\'a}r, and Ross Girshick.
\newblock Mask r-cnn.
\newblock In \emph{ICCV}, 2017{\natexlab{a}}.

\bibitem[He et~al.(2017{\natexlab{b}})He, Gkioxari, Doll{\'a}r, and Girshick]{maskrcnn}
K. He, G. Gkioxari, P. Doll{\'a}r, and R. Girshick.
\newblock Mask r-cnn.
\newblock In \emph{ICCV}, 2017{\natexlab{b}}.

\bibitem[He et~al.(2022)He, Chen, Xie, Li, Doll{\'a}r, and Girshick]{he2022masked}
K. He, X. Chen, S. Xie, Y. Li, P. Doll{\'a}r, and R. Girshick.
\newblock Masked autoencoders are scalable vision learners.
\newblock In \emph{CVPR}, 2022.

\bibitem[Hu et~al.(2022)Hu, Shen, Wallis, Allen-Zhu, Li, Wang, Wang, and Chen]{hu2021lora}
E.~J. Hu, Y. Shen, P. Wallis, Z. Allen-Zhu, Y. Li, S. Wang, L. Wang, and W. Chen.
\newblock Lora: Low-rank adaptation of large language models.
\newblock In \emph{ICLR}, 2022.

\bibitem[Jia et~al.(2021)Jia, Yang, Xia, Chen, Parekh, Pham, Le, Sung, Li, and Duerig]{align}
C. Jia, Y. Yang, Y. Xia, Y.-T. Chen, Z. Parekh, H. Pham, Q. Le, Y.-H. Sung, Z. Li, and T. Duerig.
\newblock Scaling up visual and vision-language representation learning with noisy text supervision.
\newblock In \emph{ICML}, 2021.

\bibitem[Jia et~al.(2022)Jia, Tang, Chen, Cardie, Belongie, Hariharan, and Lim]{jia2022visual}
M. Jia, L. Tang, B.-C. Chen, C. Cardie, S. Belongie, B. Hariharan, and S.-N. Lim.
\newblock Visual prompt tuning.
\newblock In \emph{ECCV}, 2022.

\bibitem[Jiao et~al.(2024)Jiao, Zhu, Huang, Zhao, Wei, and Humphrey]{maft+}
Siyu Jiao, Hongguang Zhu, Jiannan Huang, Yao Zhao, Yunchao Wei, and Shi Humphrey.
\newblock Collaborative vision-text representation optimizing for open-vocabulary segmentation.
\newblock In \emph{European Conference on Computer Vision}, 2024.

\bibitem[Kirillov et~al.(2019)Kirillov, He, Girshick, Rother, and Doll{\'a}r]{panoptic}
A. Kirillov, K. He, R. Girshick, C. Rother, and P. Doll{\'a}r.
\newblock Panoptic segmentation.
\newblock In \emph{CVPR}, 2019.

\bibitem[Li et~al.(2022{\natexlab{a}})Li, Weinberger, Belongie, Koltun, and Ranftl]{li2022languagedriven}
B. Li, K.~Q Weinberger, S. Belongie, V. Koltun, and R. Ranftl.
\newblock Language-driven semantic segmentation.
\newblock In \emph{ICLR}, 2022{\natexlab{a}}.

\bibitem[Li and Liang(2021)]{li2021prefix}
X. Li and P. Liang.
\newblock Prefix-tuning: Optimizing continuous prompts for generation.
\newblock In \emph{ACL}, 2021.

\bibitem[Li et~al.(2022{\natexlab{b}})Li, Mao, Girshick, and He]{li2022exploring}
Y. Li, H. Mao, R. Girshick, and K. He.
\newblock Exploring plain vision transformer backbones for object detection.
\newblock In \emph{ECCV}, 2022{\natexlab{b}}.

\bibitem[Liang et~al.(2023)Liang, Wu, Dai, Li, Zhao, Zhang, Zhang, Vajda, and Marculescu]{liang2023open}
F. Liang, B. Wu, X. Dai, K. Li, Y. Zhao, H. Zhang, P. Zhang, P. Vajda, and D. Marculescu.
\newblock Open-vocabulary semantic segmentation with mask-adapted clip.
\newblock In \emph{CVPR}, 2023.

\bibitem[Lin et~al.(2014)Lin, Maire, Belongie, Hays, Perona, Ramanan, Doll{\'a}r, and Zitnick]{coco}
T. Lin, M. Maire, S. Belongie, J. Hays, P. Perona, D. Ramanan, P. Doll{\'a}r, and C.~L. Zitnick.
\newblock Microsoft coco: Common objects in context.
\newblock In \emph{ECCV}, 2014.

\bibitem[Liu et~al.(2024)Liu, Li, Wu, and Lee]{liu2024visual}
H. Liu, C. Li, Q. Wu, and Y.~J. Lee.
\newblock Visual instruction tuning.
\newblock In \emph{NeurIPS}, 2024.

\bibitem[Long et~al.(2015)Long, Shelhamer, and Darrell]{fcn}
J. Long, E. Shelhamer, and T. Darrell.
\newblock Fully convolutional networks for semantic segmentation.
\newblock In \emph{CVPR}, 2015.

\bibitem[Loshchilov and Hutter(2019)]{loshchilov2017decoupled}
I. Loshchilov and F. Hutter.
\newblock Decoupled weight decay regularization.
\newblock In \emph{ICLR}, 2019.

\bibitem[Mottaghi et~al.(2014)Mottaghi, Chen, Liu, Cho, Lee, Fidler, Urtasun, and Yuille]{mottaghi_cvpr14}
R. Mottaghi, X. Chen, X. Liu, N.-G. Cho, S.-W. Lee, S. Fidler, R. Urtasun, and A. Yuille.
\newblock The role of context for object detection and semantic segmentation in the wild.
\newblock In \emph{CVPR}, 2014.

\bibitem[Neuhold et~al.(2017)Neuhold, Ollmann, Rota~Bulo, and Kontschieder]{mapillary}
Gerhard Neuhold, Tobias Ollmann, Samuel Rota~Bulo, and Peter Kontschieder.
\newblock The mapillary vistas dataset for semantic understanding of street scenes.
\newblock In \emph{ICCV}, 2017.

\bibitem[Pate et~al.(2021)Pate, Xu, Yang, Love, Ganguri, and Wong]{robot1}
S. Pate, W. Xu, Z. Yang, M. Love, S. Ganguri, and L.~L. Wong.
\newblock Natural language for human-robot collaboration: Problems beyond language grounding.
\newblock \emph{arXiv:2110.04441}, 2021.

\bibitem[Qin et~al.(2023)Qin, Wu, Yan, Li, Yuxi, Xiao, Wang, Wang, Wen, Pan, et~al.]{qin2023freeseg}
J. Qin, J. Wu, P. Yan, M. Li, R. Yuxi, X. Xiao, Y. Wang, R. Wang, S. Wen, X. Pan, et~al.
\newblock Freeseg: Unified, universal and open-vocabulary image segmentation.
\newblock In \emph{CVPR}, 2023.

\bibitem[Radford et~al.(2021)Radford, Kim, Hallacy, Ramesh, Goh, Agarwal, Sastry, Askell, Mishkin, Clark, Krueger, and Sutskever]{clip}
A. Radford, J.~W. Kim, C. Hallacy, A. Ramesh, G. Goh, S. Agarwal, G. Sastry, A. Askell, P. Mishkin, J. Clark, G. Krueger, and I. Sutskever.
\newblock Learning transferable visual models from natural language supervision.
\newblock In \emph{ICML}, 2021.

\bibitem[Rombach et~al.(2022)Rombach, Blattmann, Lorenz, Esser, and Ommer]{sd}
Robin Rombach, Andreas Blattmann, Dominik Lorenz, Patrick Esser, and Bj{\"o}rn Ommer.
\newblock High-resolution image synthesis with latent diffusion models.
\newblock In \emph{CVPR}, 2022.

\bibitem[Shalev-Shwartz and Ben-David(2014)]{shalev2014understanding}
S. Shalev-Shwartz and S. Ben-David.
\newblock Understanding machine learning: From theory to algorithms, 2014.

\bibitem[Sun et~al.(2023)Sun, Fang, Wu, Wang, and Cao]{EVA-CLIP}
Q. Sun, Y. Fang, L. Wu, X. Wang, and Y. Cao.
\newblock Eva-clip: Improved training techniques for clip at scale.
\newblock \emph{arXiv:2303.15389}, 2023.

\bibitem[Toromanoff et~al.(2020)Toromanoff, Wirbel, and Moutarde]{toromanoff2020end}
M. Toromanoff, E. Wirbel, and F. Moutarde.
\newblock End-to-end model-free reinforcement learning for urban driving using implicit affordances.
\newblock In \emph{CVPR}, 2020.

\bibitem[Wu et~al.(2024)Wu, Li, Xu, Yuan, Ding, Yang, Li, Zhang, Tong, Jiang, et~al.]{wu2024towards}
J. Wu, X. Li, S. Xu, H. Yuan, H. Ding, Y. Yang, X. Li, J. Zhang, Y. Tong, X. Jiang, et~al.
\newblock Towards open vocabulary learning: A survey.
\newblock \emph{PAMI}, 2024.

\bibitem[Wu et~al.(2019)Wu, Kirillov, Massa, Lo, and Girshick]{wu2019detectron2}
Yuxin Wu, Alexander Kirillov, Francisco Massa, Wan-Yen Lo, and Ross Girshick.
\newblock Detectron2.
\newblock \url{https://github.com/facebookresearch/detectron2}, 2019.

\bibitem[Xian et~al.(2018)Xian, Lampert, Schiele, and Akata]{xian2018zero}
Y. Xian, C.~H. Lampert, B. Schiele, and Z. Akata.
\newblock Zero-shot learning—a comprehensive evaluation of the good, the bad and the ugly.
\newblock \emph{PAMI}, 2018.

\bibitem[Xie et~al.(2024)Xie, Cao, Xie, Khan, and Pang]{sed}
B. Xie, J. Cao, J. Xie, F.~S. Khan, and Y. Pang.
\newblock Sed: A simple encoder-decoder for open-vocabulary semantic segmentation.
\newblock In \emph{CVPR}, 2024.

\bibitem[Xu et~al.(2023{\natexlab{a}})Xu, Liu, Vahdat, Byeon, Wang, and Mello]{odise}
J. Xu, S. Liu, A. Vahdat, W. Byeon, X. Wang, and S.~De Mello.
\newblock Open-vocabulary panoptic segmentation with text-to-image diffusion models.
\newblock In \emph{CVPR}, 2023{\natexlab{a}}.

\bibitem[Xu et~al.(2022)Xu, Zhang, Wei, Lin, Cao, Hu, and Bai]{xu2022simple}
M. Xu, Z. Zhang, F. Wei, Y. Lin, Y. Cao, H. Hu, and X. Bai.
\newblock A simple baseline for open-vocabulary semantic segmentation with pre-trained vision-language model.
\newblock In \emph{ECCV}, 2022.

\bibitem[Xu et~al.(2023{\natexlab{b}})Xu, Zhang, Wei, Hu, and Bai]{san}
M. Xu, Z. Zhang, F. Wei, H. Hu, and X. Bai.
\newblock Side adapter network for open-vocabulary semantic segmentation.
\newblock In \emph{CVPR}, 2023{\natexlab{b}}.

\bibitem[Xu et~al.(2023{\natexlab{c}})Xu, Xiong, Ding, and Tu]{masqclip}
X. Xu, T. Xiong, Z. Ding, and Z. Tu.
\newblock Masqclip for open-vocabulary universal image segmentation.
\newblock In \emph{ICCV}, 2023{\natexlab{c}}.

\bibitem[Yu et~al.(2023)Yu, He, Deng, Shen, and Chen]{fcclip}
Q. Yu, J. He, X. Deng, X. Shen, and L.-C. Chen.
\newblock Convolutions die hard: Open-vocabulary segmentation with single frozen convolutional clip.
\newblock In \emph{NeurIPS}, 2023.

\bibitem[Zhang et~al.(2024)Zhang, Han, Zhou, Hu, Yan, Lu, Li, Gao, and Qiao]{zhang2023llama}
R. Zhang, J. Han, A. Zhou, X. Hu, S. Yan, P. Lu, H. Li, P. Gao, and Y. Qiao.
\newblock Llama-adapter: Efficient fine-tuning of language models with zero-init attention.
\newblock In \emph{ICLR}, 2024.

\bibitem[Zhao et~al.(2024)Zhao, Tu, Wei, Mei, and Xie]{zhao2023tuning}
B. Zhao, H. Tu, C. Wei, J. Mei, and C. Xie.
\newblock Tuning layernorm in attention: Towards efficient multi-modal llm finetuning.
\newblock In \emph{ICLR}, 2024.

\bibitem[Zhou et~al.(2019)Zhou, Zhao, Puig, Xiao, Fidler, Barriuso, and Torralba]{zhou2019semantic}
B. Zhou, H. Zhao, X. Puig, T. Xiao, S. Fidler, A. Barriuso, and A. Torralba.
\newblock Semantic understanding of scenes through the ade20k dataset.
\newblock \emph{IJCV}, 2019.

\bibitem[Zhu and Chen(2023)]{zhu2023survey}
C. Zhu and L. Chen.
\newblock A survey on open-vocabulary detection and segmentation: Past, present, and future.
\newblock \emph{PAMI}, 2023.

\end{thebibliography}
